\documentclass[twoside]{article}

\usepackage[accepted]{aistats2019}
\usepackage[capitalize]{cleveref}
\usepackage{graphicx,tikz}
\usepackage{booktabs}
\usepackage{amssymb}
\usepackage{caption,subcaption}
\usepackage{multirow,multicol,bigdelim}
\usepackage{algorithm2e}
\usepackage{mlmacros}
\usepackage{cancel}

\crefname{algocf}{alg.}{algs.}
\Crefname{algocf}{Algorithm}{Algorithms}

\usetikzlibrary{decorations.pathreplacing,calc}

\tikzset{brace/.style={decorate, decoration={brace}},
 brace mirrored/.style={decorate, decoration={brace,mirror}},
}

\newcounter{brace}
\probdists{p,q}
\DeclareMathOperator*{\argmax}{arg\,max}
\DeclareMathOperator*{\argmin}{arg\,min}
\newcommand*\xbar[1]{%
  \hbox{%
    \vbox{%
      \hrule height 0.5pt % The actual bar
      \kern0.5ex%         % Distance between bar and symbol
      \hbox{%
        \kern-0.3em%      % Shortening on the left side
        \ensuremath{#1}%
        \kern-0.1em%      % Shortening on the right side
      }%
    }%
  }%
} 
\MkProbDist{TN}{\xbar{\mathcal{N}}}
\MkProbDist{N}{\mathcal{N}}
\begin{document}

% If your paper is accepted and the title of your paper is very long,
% the style will print as headings an error message. Use the following
% command to supply a shorter title of your paper so that it can be
% used as headings.
%
%\runningtitle{I use this title instead because the last one was very long}

% If your paper is accepted and the number of authors is large, the
% style will print as headings an error message. Use the following
% command to supply a shorter version of the authors names so that
% they can be used as headings (for example, use only the surnames)
%
%\runningauthor{Surname 1, Surname 2, Surname 3, ...., Surname n}

\twocolumn[

\aistatstitle{Bayesian Learning of Neural Network Architectures}

\aistatsauthor{ Georgi Dikov \And Patrick van der Smagt \And Justin Bayer }

\aistatsaddress{ bayerj@argmax.ai \\ AI Research, Volkswagen Group, Munich, Germany } ]

\begin{abstract}
  In this paper we propose a Bayesian method for estimating architectural parameters of neural networks, namely layer size and network depth. 
  We do this by learning concrete distributions over these parameters. 
  Our results show that regular networks with a learnt structure can generalise better on small datasets, while fully stochastic networks can be more robust to parameter initialisation. 
  The proposed method relies on standard neural variational learning and, unlike randomised architecture search, does not require a retraining of the model, thus keeping the computational overhead at minimum.
\end{abstract}

\section{INTRODUCTION}

One of the reasons for the success of modern deep learning models is attributed to the development of powerful architectures that exploit certain regularities in the data (e.g., convolutional networks such as~\cite{simonyan2014very, szegedy2015going}) and alleviate issues with numerical optimisation (e.g., learning an identity mapping in very deep networks~\cite{he2016identity}).
In fact, it has been shown \cite{saxe2011random} that architecture alone can improve representation learning even with randomly initialised weights.

Traditionally, the architecture of a neural network is treated as a set of static hyperparameters, which are tuned based on an observed performance on a held-out validation set.
This viewpoint, however, requires that a network is initialised, trained until convergence and evaluated at each modification of the architecture---a time-consuming procedure which does not allow for an efficient, exhaustive hyperparameter search.

In this work, we propose a scalable Bayesian method to structure optimisation by treating hyperparameters, such as the layer size and network depth, as random variables whose parameterised distributions are learnt together with the rest of the network weights. 
Taking a Bayesian probabilistic approach to architecture learning is good for two main reasons: (i)~the posterior distribution over the architectural parameters reveals whether or not the model has the capacity to represent the training data well; and~(ii) imposing prior beliefs over the parameters naturally allows for expert knowledge to be incorporated into the model, without imposing any unbreakable constraints as a side effect.
However, obtaining the correct posterior distribution in closed form is not possible due to the highly nonlinear nature of deep neural networks; also residing to a Markov Chain Monte Carlo sampling technique is computationally prohibitive.
Instead, we apply the framework of approximate variational inference in order to estimate a posterior distribution over the architectural variables and maintain the differentiability of the model by the means of a continuous relaxation on the discrete categorical (concrete) distribution~\cite{maddison2016concrete, jang2016categorical}.
Thus we are able to efficiently evaluate a continuum of architectures.
We will show empirically that ensembling predictions from networks of sampled architectures acts as a regulariser and mitigates overfitting.

In the next section we review the necessary background in approximate variational inference, present our model from a Bayesian viewpoint and briefly introduce the concrete categorical distribution.
In~\cref{sec:adaptive_arch} we show the mechanism of layer size and network depth learning and give an intuitive interpretation of the approach.
\Cref{sec:related_work} compares our method to existing ones and discusses their shortcomings.
In~\cref{sec:experiments} we evaluate multiple models in regression, classification and bandits tasks and finally we discuss potential consequences in~\cref{sec:conclusion}.

\section{BACKGROUND AND MODEL STATEMENT}
\label{sec:background}
 
\subsection{Approximate Variational Inference}
Let $\mathbf{W}=\{\mathbf{W}^1, \mathbf{W}^2, \dots, \mathbf{W}^n\}$ denote the weights of an $n$-layer network and $\boldsymbol{\alpha}$ the architectural parameters which are going to be learnt.
Further, let $(\mathbf{X}, \mathbf{Y})$ be a labelled dataset.
Then, in the framework of Bayesian reasoning, we define a prior distribution $\p{\mathbf{W}, \boldsymbol{\alpha}} = \p{\mathbf{W}}\p{\boldsymbol{\alpha}}$, a likelihood model $\p{\mathbf{Y}}{\mathbf{X}, \mathbf{W}, \boldsymbol{\alpha}}$ and we seek to infer the posterior distribution $\p{\mathbf{W}, \boldsymbol{\alpha}}{\mathbf{X}, \mathbf{Y}}$.
The latter, however, cannot be evaluated precisely due to the intractability of the normalisation constant $\p{\mathbf{Y}}{\mathbf{X}}$.
The variational Bayes approach reframes the problem of inferring the posterior distribution into an optimisation one, by minimising an approximation error between a parameterised surrogate distribution $\q{\mathbf{W}, \boldsymbol{\alpha}}{\mathbf{X}, \mathbf{Y}}$ and the posterior distribution.
For the sake of computational simplicity, throughout this work we will assume that the approximate posterior is fully factorisable, \ie:
\begin{equation}
    \q{\mathbf{W}, \boldsymbol{\alpha}}{\mathbf{X}, \mathbf{Y}} = \prod_{l=1}^n \q{\mathbf{W}^l}{\mathbf{X},  \mathbf{Y}}
    \prod_{\alpha \in \boldsymbol{\alpha}} \q{\alpha}{\mathbf{X}, \mathbf{Y}}
\label{eq:factorisation}
\end{equation}
and that the network weights in each layer $l$, $\mathbf{W}^l$, are independent and Gaussian distributed with parameters $\boldsymbol{\mu}, \boldsymbol{\sigma} \in \mathbb{R}^{|\mathbf{W}^l|}$, \ie
$\q{\mathbf{W}^l}{\mathbf{X}, \mathbf{Y}} = \N{\mathbf{W}^l}{\boldsymbol{\mu},\, \mathrm{diag}(\boldsymbol{\sigma}^2)}$.
Note that relaxing the independence and/or the functional form assumption on the network weights can improve modelling performance, as shown by~\cite{cremer2018inference,pawlowski2017implicit}.
Nevertheless, we leave the extension of architecture learning in Bayesian neural networks with more sophisticated posterior approximation to future work.
The prior distribution over the weights $\mathbf{W}^l$ will be a zero-mean factorised Gaussian with the same fixed variance $\sigma_0^2$ for each weight, \ie $\p{\mathbf{W}^l} = \N{\mathbf{W}^l}{\mathbf{0},\, \sigma_0^2 I}$.

The specific form of $\q{\boldsymbol{\alpha}}$ and $\p{\boldsymbol{\alpha}}$ will be elaborated in detail in~\cref{sec:adaptive_arch} where we will consider learning the layer sizes and the overall network depth.
Due to the discrete nature of these parameters, we cannot use backpropagation to learn their posteriors. We will show in~\cref{sec:reparam_trick,sec:concrete_dist} how we could circumvent this issue.

Let $\boldsymbol{\eta}$ and $\boldsymbol{\theta}$ represent the sets of variational parameters for the approximate marginals $\q[\boldsymbol{\eta}]{\mathbf{W}}{\mathbf{X}, \mathbf{Y}}$ and $\q[\boldsymbol{\theta}]{\boldsymbol{\alpha}}{\mathbf{X}, \mathbf{Y}}$ which we denote as $\q[\boldsymbol{\eta}]{\mathbf{W}}$ and $\q[\boldsymbol{\theta}]{\boldsymbol{\alpha}}$ respectively.  
One way to quantify the approximation error between the surrogate $q$ and the true posterior $p$ is to measure their Kullback-Leibler divergence~\cite{kullback1951information}.
It can be shown that the following relation holds~\cite{jordan1999introduction}:
\begin{align}
\boldsymbol{\eta}^*, \boldsymbol{\theta}^* 
&= \argmin_{\boldsymbol{\eta}, \boldsymbol{\theta}} \kl{\q[\boldsymbol{\eta}]{\mathbf{W}}\q[\boldsymbol{\theta}]{\boldsymbol{\alpha}}}{\p{\mathbf{W}, \boldsymbol{\alpha}}{\mathbf{X},\, \mathbf{Y}}} \\
&
\begin{aligned}
    =\argmin_{\boldsymbol{\eta}, \boldsymbol{\theta}} 
    &- \expc[\q[\boldsymbol{\eta}]{\mathbf{W}}\q[\boldsymbol{\theta}]{\boldsymbol{\alpha}}]{\log{\p{\mathbf{Y}}{\mathbf{X}, \mathbf{W}, \boldsymbol{\alpha}}}} \\
    &+ \kl{\q[\boldsymbol{\eta}]{\mathbf{W}}}{\p{\mathbf{W}}} \\
    &+ \kl{\q[\boldsymbol{\theta}]{\boldsymbol{\alpha}}}{\p{\boldsymbol{\alpha}}}
\end{aligned} \\
&=\argmin_{\boldsymbol{\eta}, \boldsymbol{\theta}}-\mathcal{L}_\text{ELBO}(\boldsymbol{\eta}, \boldsymbol{\theta}, \mathbf{X}, \mathbf{Y}).
\label{eq:elbo}
\end{align}
The quantity in \cref{eq:elbo}, $\mathcal{L}_\text{ELBO}$, is called the \textit{Evidence Lower Bound} and will be approximated with Monte Carlo (MC) sampling since the prior, the approximate posterior and the likelihood distributions will have known densities as we will see in~\cref{sec:adaptive_arch}. 
Also, given that the prior distribution $\p{\mathbf{W}}$ is a Gaussian, the KL-divergence term for the network weights will be computed analytically and thus will reduce the variance in the gradient estimates. 
However, the KL-divergence for the architectural parameters $\boldsymbol{\alpha}$ will be estimated using MC sampling.
Finally, using the approximations $\q[\boldsymbol{\eta}]{\mathbf{W}}$ and $\q[\boldsymbol{\theta}]{\boldsymbol{\alpha}}$ we can define a posterior predictive distribution over the labels $\mathbf{Y}$ and approximate it with MC sampling:
\begin{align}
    \p{\mathbf{Y}}{\mathbf{X}} = \iint \p{\mathbf{Y}}{\mathbf{X}, \mathbf{W}, \boldsymbol{\alpha}}  \q[\boldsymbol{\eta}]{\mathbf{W}}
    \q[\boldsymbol{\theta}]{\boldsymbol{\alpha}}
    d\mathbf{W} d\boldsymbol{\alpha}.
\end{align}
Note that even if we treat the network weights $\mathbf{W}$ as point estimates we can still compute an approximate posterior distribution over $\boldsymbol{\alpha}$ and optimise it using the ELBO objective while performing a MAP estimate over $\mathbf{W}$. 
That is, the approach of Bayesian architecture learning is applicable to regular neural networks as well and we will show such an example in~\cref{sec:experiments}.

\subsection{The Reparameterisation Trick}
\label{sec:reparam_trick}
The reparameterisation trick \cite{kingma2013auto} refers to a technique of representing sampling from a probability distribution as a deterministic operation over the distributional parameters and an external source of independent noise. 
In the context of architecture learning we would like to show that such a reparameterisation is possible for the architectural random variable $\boldsymbol{\alpha}$ of some $\boldsymbol{\theta}$-parameterised distribution $\boldsymbol{\alpha} \sim \q[\boldsymbol{\theta}]{\boldsymbol{\alpha}}$. 
Then, if there is a deterministic and differentiable function $g$ such that $\boldsymbol{\alpha} = g(\boldsymbol{\theta}, \boldsymbol{\epsilon})$ with $\boldsymbol{\epsilon} \sim \p{\boldsymbol{\epsilon}}$ guaranteeing that $\expc[\q[\boldsymbol{\theta}]{\boldsymbol{\alpha}}]{\boldsymbol{\alpha}} = \expc[\p{\boldsymbol{\epsilon}}]{g(\boldsymbol{\theta}, \boldsymbol{\epsilon})}$, we can compute the gradient \wrt $\boldsymbol{\theta}$ on $g$ and use standard backpropagation to learn $\boldsymbol{\theta}$.

\subsection{The Concrete Categorical Distribution}
\label{sec:concrete_dist}
Proposed by~\cite{jang2016categorical,maddison2016concrete} the Gumbel-softmax or concrete categorical distribution is a continuous extension of its discrete counterpart.
It is fully reparameterisable as sampling $K$-dimensional probability vectors $\mathbf{s} \in \Delta^{K-1}$ can be expressed as a deterministic function of its parameters---the probability vector $\boldsymbol{\pi}$---and an external source of randomness $\boldsymbol{\epsilon}$ which is Gumbel-distributed:
\begin{align*}
s_i &= \frac{\exp((\log\pi_i + \epsilon_i)/\tau)}{\sum_j\exp((\log\pi_j + \epsilon_j)/\tau)}, \\ \epsilon_i &\sim -\log(-\log\bigl(\text{Uniform}(0, 1))\bigr).
\label{eq:reparam_sample_conccat}
\end{align*}
Here $\tau$ is a temperature hyperparameter controlling the smoothness of the approximation. For $\tau \rightarrow 0$ the samples become one-hot vectors and for $\tau \rightarrow \infty: \, s_i~=~s_j, \,  \forall i,j$. In this work we will consider $\tau$ fixed.
The density of the concrete categorical distribution is
\begin{equation}
    \p{\mathbf{s}}{\boldsymbol{\pi}, \tau} = (K-1)!\tau^{K-1}\frac{\prod_{i = 1}^K \pi_i s_i^{-\tau-1}}{\left(\sum_{i = 1}^K \pi_i s_i^{-\tau}\right)^K}.
\label{eq:pdf_concrete_categorical}
\end{equation}
Analogously for the binary case ($s \in [0, 1]$), one can express a sample from a concrete Bernoulli distribution by perturbing the logit with noise from a Logistic distribution and squashing it through a sigmoid:
\begin{align*}
s &= \frac{1}{1 + \exp(-(\log(\pi) - \log(1 - \pi) + \epsilon)/\tau)}, \\ 
\epsilon &\sim \mathrm{Logistic}(0, 1).
\end{align*}
The functional form of its density function is given as:
\begin{equation}
    \p{s}{\pi, \tau} = \frac{\tau\pi s^{-\tau - 1}(1 - s)^{-\tau-1}}{(\pi s^{-\tau} + (1 - s)^{-\tau})^2}.
\label{eq:concrete_bernoulli_density}
\end{equation}
For more properties of the concrete distributions see the appendices in~\cite{jang2016categorical,maddison2016concrete}.

\section{ADAPTIVE NETWORK ARCHITECTURE}
\label{sec:adaptive_arch}
In this work we will focus on two important architectural hyperparameters but analogous extensions to others are possible. 
First we will look into learning the size of an arbitrary layer $l$ denoted with $\mathbf{s}^l$ and then we will proceed with estimating the optimal depth of a network by means of independent layer-wise skip connections $\gamma^l$. 
Following the independence assumption from~\cref{eq:factorisation} for a network of $n$ layers we have: 
\begin{equation}
    \q{\boldsymbol{\alpha}} = \prod_{l=1}^n \q{\mathbf{s}^l}\q{\gamma^l}.
\end{equation}
Analogous factorisation applies for the prior $\p{\boldsymbol{\alpha}}$ as well. In our work, it has the same functional form as the approximate posterior but has fixed parameters.

\subsection{Layer Size}

Let $\mathbf{s}^l \in \Delta^{K-1}$ be a concrete-categorically distributed random variable encoding the size of an arbitrary fully-connected layer $l$ with maximum capacity of $K$ units\footnote{Or $K$ filters if the layer is convolutional.}. 
Then the integer number representing the layer size encoded in a sample is given as $k = \argmax_i s^l_i$. 
In order to enforce the sampled size on the layer, we propose building a soft and differentiable mask $m(\mathbf{s}^l) \in \Delta^{K-1}$ which multiplicatively gates the output of $l$: 
\begin{equation}
\mathbf{y}^l = f(\mathbf{W}^l \mathbf{y}^{l-1}) \odot m(\mathbf{s}^l)
\label{eq:adasize}
\end{equation}
where we omit the bias $\mathbf{b}^l$ for the sake of notational brevity and use $f$ to denote the activation function.
Due to the fully-connected nature of the layer, there is in general no preference for which $k$ units should be used. 
However, one has to be consistent in selecting them across different gradient updates, as this subset of units will represent the reduced in size layer and all others should be discarded, e.g.\ by deleting $K - k$ rows of $\mathbf{W}^l$.
To do this, we construct the mask such that the top $k$ rows are approximately 1s (letting through gradient updates) and the rest 0s (blocking gradient updates). \Eg $m(\mathbf{s}^l) = \mathbf{U}\mathbf{s}^l$ where $\mathbf{U} \in \{0, 1\}^{K \times K}$ is an upper triangular matrix of ones. Since $\mathbf{s}^l$ will never be a one-hot vector in practice, the resulting mask will be soft.
Note that in a fully Bayesian neural network, the approximate posterior on the parameters of all redundant (blocked) units will conform to the prior, essentially paying a portion of the divergence debt borrowed by the active units.

Before giving explicitly the form of the approximate posterior $\q{\mathbf{s}^l}$ we argue that (i)~the learnt distribution should be unimodal, such that a unique optimal layer size can be deduced, and (ii)~it should provide us with a meaningful uncertainty estimate. 
As the probabilities of the concrete categorical distribution are not constrained to express unimodality, we suggest to limit the degrees of freedom by coupling $\boldsymbol{\pi}_i$ through a deterministic and differentiable function. 
One such candidate is the renormalised density of the truncated Normal distribution which we denote as $\,\TN{\mu, \sigma^2, 1, K}$. 
By abuse of notation we express $\boldsymbol{\pi}$ as a function of $\mu$ and $\sigma$ and evaluate it at points $\{1, 2, \dots, K\}$:
\begin{align}
    \boldsymbol{\pi}(\mu,\, \sigma)_i &= \frac{\TN{i}{\mu,\, \sigma^2,\, 1,\, K}}{\sum_{j=1}^K\TN{j}{\mu,\, \sigma^2,\, 1,\, K}} \quad \text{ for $i \in [K]$} \label{eq:adasize_reparam},\\
    \q[\mu, \, \sigma]{\mathbf{s}^l} &= \mathrm{ConcreteCategorical}(\boldsymbol{\pi}(\mu,\, \sigma)).
\label{eq:adasize_reparam2}
\end{align}
Besides the unimodality, this parameterisation is also advantageous for requiring a constant number of variational parameters \wrt the layer size. 
Throughout this work, the prior $\p[\mu_0, \sigma_0]{\mathbf{s}^l}$ assumes the same parameterisation as $\q[\mu, \sigma]{\mathbf{s}^l}$ and $\mu_0$ and $\sigma_0$ are specified in advance.
Care must be taken, however, when setting the temperature $\tau$. 
Since the gradient is scaled with the inverse of $\tau$, small values, e.g.\ in the order of $0.01$, can lead to optimisation instability. 
We have observed a good performance with a constant temperature in the range of $1.0$ to $3.0$, which we found empirically. 
Finally, we note that the gradients \wrt the weights and biases are multiplicatively stretched by the sampled mask vector. 
Therefore, our method can be interpreted as an auxiliary per-unit learning rate, modulating the error signal coming from the data log-likelihood term in the ELBO objective.
% \Cref{alg:adasize} summarises the learning procedure for Bayesian neural networks with adaptive layer size. 
% {\small
% \SetKwInput{KwDef}{Define}
% \SetKwInput{KwInit}{Initialise}
% % \RestyleAlgo{boxed}
% \RestyleAlgo{boxruled}
% %\LinesNumbered
% \begin{algorithm}
%     \KwDef{
%         $\boldsymbol{\theta} = \{(\mu_l, \sigma_l)\}_{l \in [n]}$;
%         $\p{\mathbf{y}}{\mathbf{x}, \mathbf{W}, \boldsymbol{\theta}}$; 
%         $\p{\boldsymbol{\mathbf{s}}}$\;
%     }
%     \KwInit{$\boldsymbol{\eta}_0 \leftarrow \boldsymbol{\theta}_\mathrm{prior}$}
%     \For{t = 1, 2 \dots}{
%         Sample $\{\mathbf{W}_1, \dots, \mathbf{W}_M\}$ from $\q[\boldsymbol{\eta}]{\mathbf{W}}$\;
%         Sample $\{\mathbf{s}_1, \dots, \mathbf{s}_M\}$ from $\q[\boldsymbol{\mu}, \boldsymbol{\sigma}]{\mathbf{s}}$\;
%         Sample $\{\boldsymbol{\gamma}_1, \dots, \boldsymbol{\gamma}_M\}$ from $\q[\pi]{\mathbf{s}}$\;
%         \For{i = 1, 2, \dots, M}{
%             $\mathbf{W}_i \sim \q[\boldsymbol{\eta}]{\mathbf{W}}$\;
%             $\mathbf{s}_i \sim \q[\boldsymbol{\theta}]{\mathbf{s}}$\;
%         }
%         $\begin{aligned}
%             \mathcal{L}_{\text{ELBO}} \approx \frac{1}{M}\sum_{i=1}^M &\log{\p{\mathbf{Y}}{\mathbf{X}, \mathbf{W}_i, \mathbf{s}_i}} \\
%             &- \log{\q[\boldsymbol{\theta}]{\mathbf{s}_i}} + \log{\p{\mathbf{s}_i}}
%         \end{aligned}$
%         $\boldsymbol{\theta}_t \leftarrow \boldsymbol{\theta}_{t-1} + \epsilon \nabla_{\boldsymbol{\theta}} \mathcal{L}_{\text{ELBO}}$
%     }
%  \caption{Adaptive size by backpropagation.}
%  \label{alg:adasize}
% \end{algorithm}
% }

\subsection{Network Depth}

Inspired by~\cite{he2016identity}, we infer the optimal depth of a feed-forward neural network by learning a bypass variable $\gamma^l$ for each layer independently. 
Using the notation from above, we can express the layer output $\mathbf{y}^l$ as
\begin{equation}
\mathbf{y}^l = (1 - \gamma^l) f(\mathbf{W}^l\mathbf{y}^{l-1}) + \gamma^l\mathbf{y}^{l-1}.
\label{eq:skip_output}
\end{equation}
We treat $\gamma^l$ in a Bayesian manner and assume a concrete Bernoulli distribution for the form of the approximate posterior. 
Thus we learn a single variational parameter $\pi$ per layer and, again, keep the temperature hyperparameter $\tau$ fixed: 
\begin{equation}
\q[\pi]{\gamma^l} = \mathrm{ConcreteBernoulli}(\gamma^l).
\label{eq:skip_posterior}
\end{equation}
We set the prior $\p[\pi_0]{\gamma^l}$ to be another concrete Bernoulli distribution with fixed parameter $\pi_0$.
Similarly to the concrete categorical distribution, the temperature hyperparameter $\tau$ cannot be small enough so that the sampled bypass coefficient $\gamma^l$ becomes a numerical 1 or 0. 
Therefore, in the process of training, the outputs of the skipped layer are only strongly inhibited and not completely shut off but as we will see, this still allows to detect an optimal layer count.

One drawback of the presented approach is its limited applicability to those layers only which do not change the dimensionality of their inputs. 
The reason is that the skip connection is implemented as a simple convex combination of the layer's input and output as given in \cref{eq:skip_output}. 
Nevertheless, this method can be used in parallel with the adaptive layer size and thus enable intermediate dimensionality fluctuations. 
Analogously to the per-unit learning rate argument, we can view the skip connection as a modulation on the gradients to all units and we interpret this method as an adaptive per-layer learning rate.
 
\section{RELATED WORK}
\label{sec:related_work}
Neural network architecture search has long been a topic of research and diverse methods such as evolutionary algorithms~\cite{todd1988evolutionary, miller1989designing, kitano1990designing}, reinforcement learning~\cite{zoph2016neural} or Bayesian optimisation~\cite{bergstra2013making, mendoza2016towards} have been applied. 
Despite the underlying differences, all these approaches share a common trait in the fact that they decouple the architecture design from the training. 
Consequently, this has a significant computational burden and to the best of our knowledge, we are the first to oppose to this paradigm and merge weight and architectural hyperparameter optimisation using the  forward- and backpropagation cycle of neural network training.

In~\cite{lecun1990optimal, hassibi1993second} unimportant weights are identified and removed from the architecture. 
A major limitation is that the initial network architecture can only be reduced. 
Our approach is similar in the sense that it has an upper limit on the network size, but it also allows for growth after initial contraction, should there be new evidence supporting it. 
Furthermore the method presented in this work is principled in the inclusion of expert knowledge in the form of fixed prior probability for each layer and only requires the manual tuning of the temperature constant $\tau$.

\section{EXPERIMENTS}
\label{sec:experiments}

\subsection{Regression on Toy Data}
\paragraph{Point-estimate Weights}

\begin{figure}[ht]
    \centering
    \includegraphics[width=0.45\textwidth]{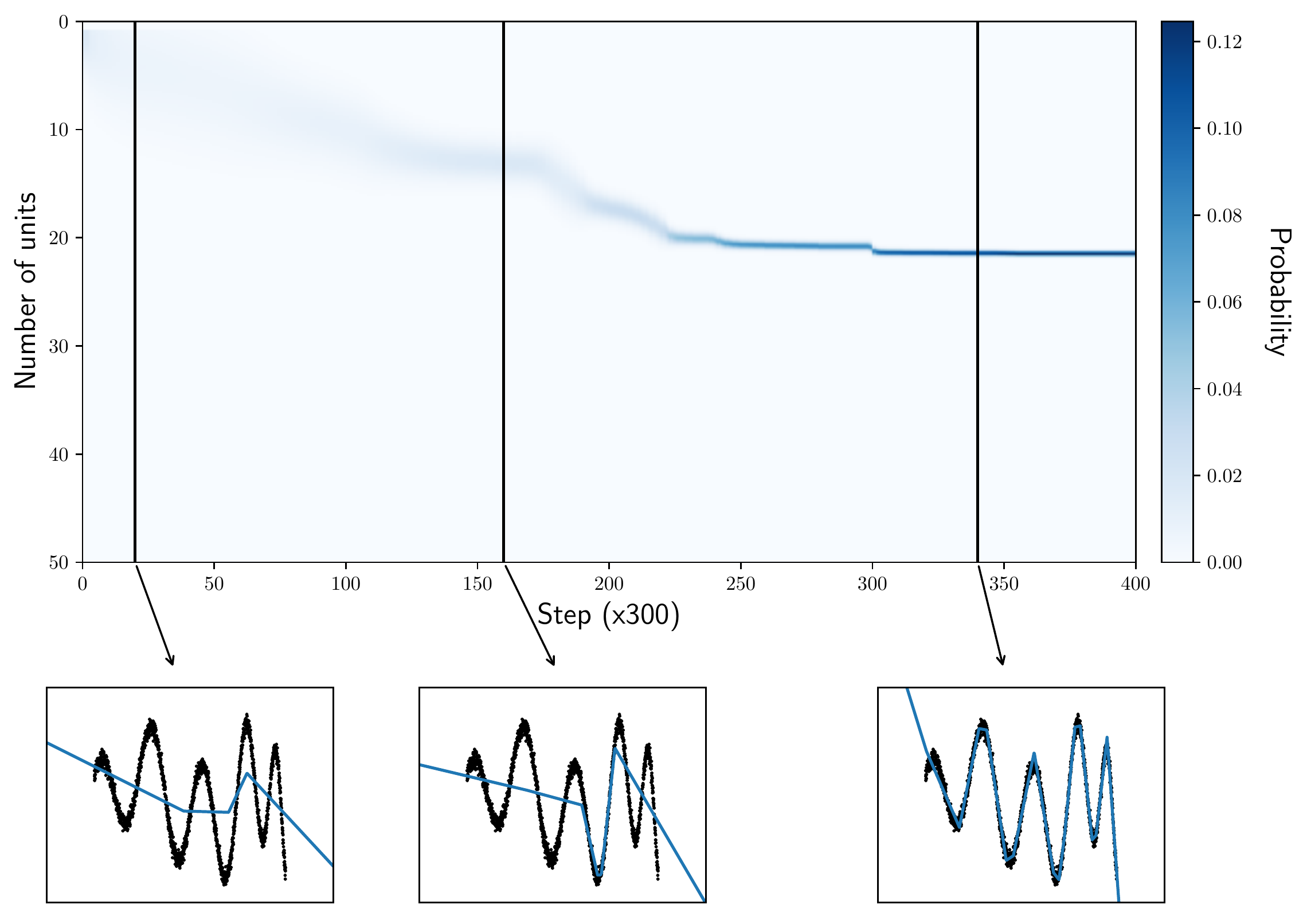}
    \caption{\small{Change in the posterior probabilities $\boldsymbol{\pi}$ over time (as used in~\cref{eq:adasize_reparam,eq:adasize_reparam2}). Below the diagram, three snapshots show the fit of the training data: the more units are released, the better the network is able to account for the non-linearity of the data. The optimisation converges to parameters $\mu=21.99$ and $\sigma=0.16$. The temperature hyperparameter $\tau$ is set to $3.0$.}}
    \label{fig:regression_adasize_pointestimate}
\end{figure}

In this first toy data experiment we demonstrate learning a suitable layer size in a single-layer neural network with 50 units and ReLU activation functions. We set a very conservative prior on the size variable $\p[\mu_0, \sigma_0]{\mathbf{s}}$ with $\mu_0=1$ and $\sigma_0=2$ and record the change in the approximate posterior over time. 
\Cref{fig:regression_adasize_pointestimate} depicts qualitatively the probabilities of the concrete categorical distribution and three snapshots show the current fit over the dataset. 

In this example, we generate 2000 points from a one-dimensional noisy periodic function. 
Due to the large number of data points, the total loss is largely dominated by the data likelihood term and the increasing divergence between the approximate posterior and the prior is acting as a weak regulariser. 
Consequently, the allocation of more units stops after the data is well approximated. 
Note that this would not happen, should the prior parameter $\mu_0$ be set to a large value, e.g.\ 40, as there is no incentive for the model to converge to a simpler solution. 
We will see in short that this is no longer the case once we treat the network weights $\mathbf{W}$ in a Bayesian way as well.

\begin{figure}[ht]
    \centering
    \includegraphics[width=0.45\textwidth]{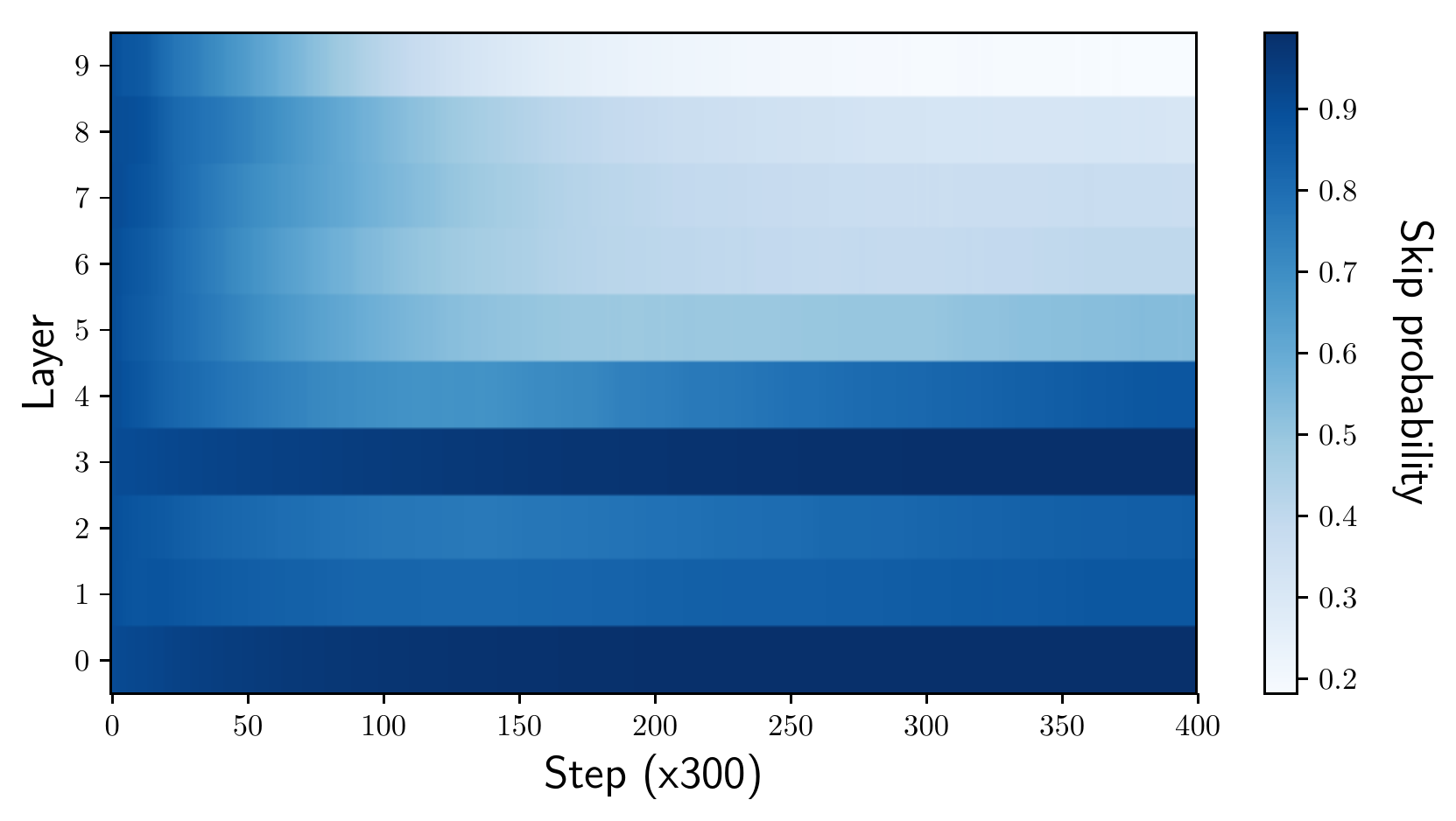}
    \caption{\small{Change in the posterior probabilities $\{\pi^1, \dots, \pi^{10}\}$ for the skip variables $\{\gamma^l\}_{l\in \{1,2,\dots,10\}}$ (see~\cref{eq:skip_posterior}) over time. Five of the layers are bypassed with high probability, indicating that a network with 5 hidden layers of 5 units each is enough to fit the data. The temperature hyperparameter $\tau$ is set to $1.0$ for each layer.}}
    \label{fig:regression_adadepth_pointestimate}
\end{figure}

Next, we initialise a deep neural network with 11 layers, 10 of which are subject to the bypassing mechanism. 
In order to enforce the usage of more than one layer we limit the size of each to 5 units and we use again a ReLU activation function. 
\Cref{fig:regression_adadepth_pointestimate} shows the change in the probability of skipping a layer over time. 
The posterior allows for a clear interpretation that a rigid network of 5 layers will be able to reliably fit the data.

\paragraph{Bayesian Weights}

\begin{figure}[h!]
\centering
	\begin{subfigure}{.24\textwidth}
		\includegraphics[width=0.9\textwidth]{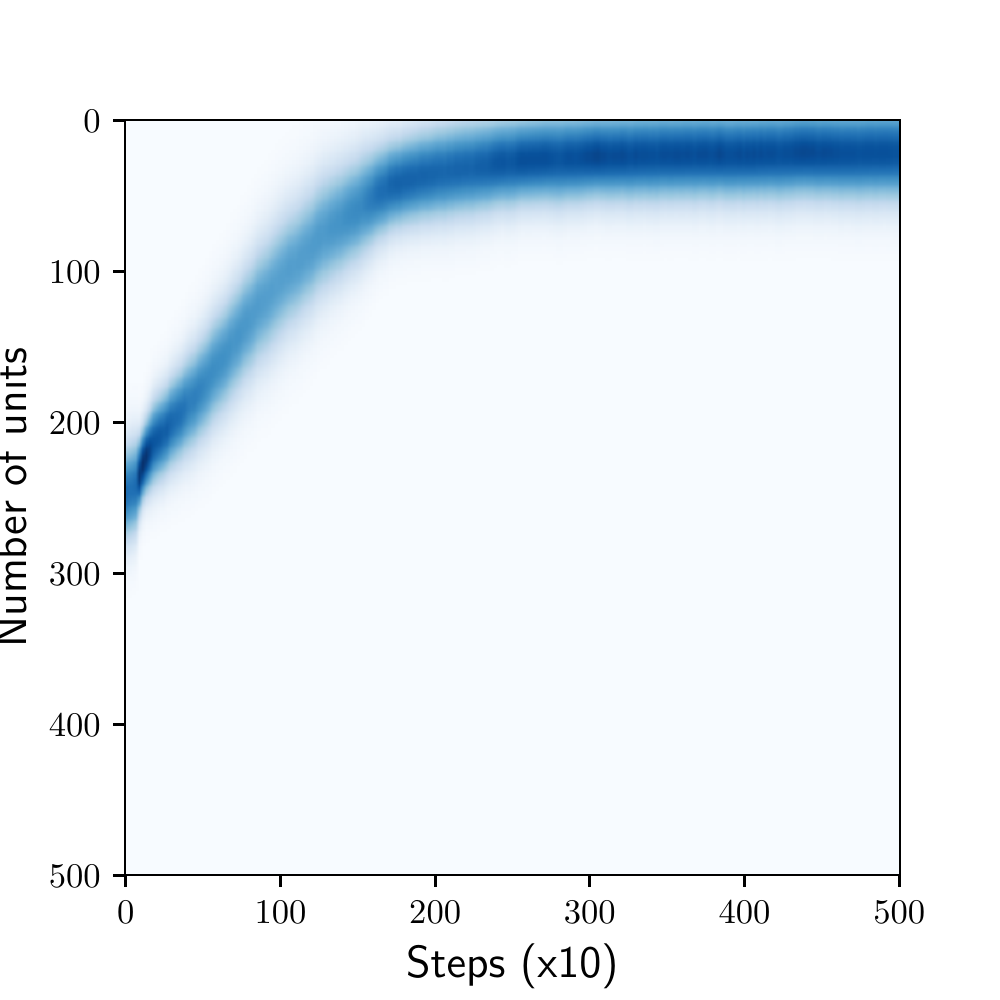}
		\caption{}
		\label{fig:regression_500units_250init}
	\end{subfigure}%
	\begin{subfigure}{.24\textwidth}
		\includegraphics[width=0.9\textwidth]{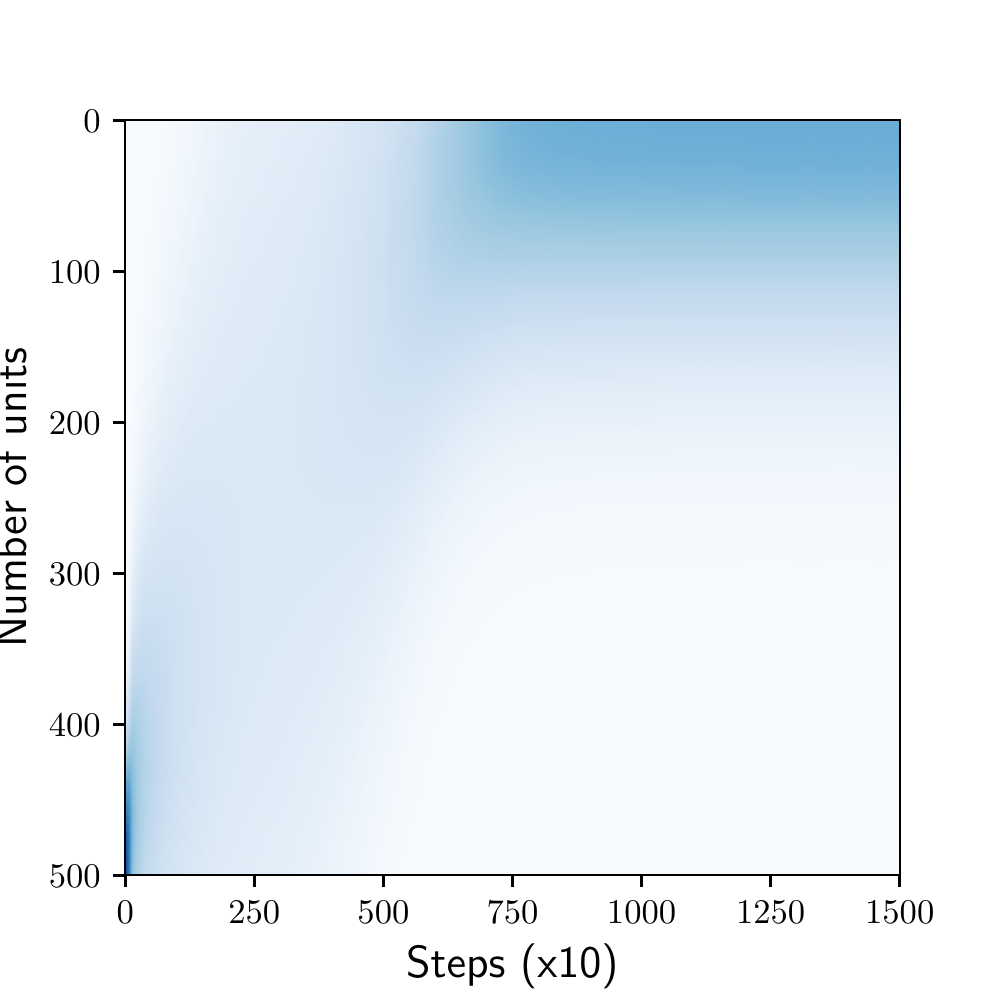}
		\caption{}
		\label{fig:regression_500units_500init}
	\end{subfigure}%
\caption{\small{Change in posterior over the size of a single-layer Bayesian neural network. Prior parameters: (a) $\mu_0=250$ and $\sigma_0=20$ and (b) $\mu_0=500$ and $\sigma_0=50$. The temperature hyperparameter $\tau$ is set to $3.0$.}}
\label{fig:regression_adasize_bbb}
\end{figure}

We now construct a fully Bayesian neural network with independently normally distributed weights and biases. 
In Bayesian neural networks the KL-divergence between the approximate posterior and the prior is acting as a strong regulariser on the parameters and in cases of small data size and overly parameterised models, the noise in the parameters dominates. 
The aim of this experiment is to show that the presented framework of architecture optimisation mitigates this issue by not only extending inadequately small architectures but also reducing oversized ones. 
\Cref{fig:regression_500units_250init,fig:regression_500units_500init} show the change in posterior for two different priors: one with $\mu_0=250$ and $\sigma_0=20$ and another with $\mu_0=500$ and $\sigma_0=50$. 
Notice that in both cases the variational parameter $\mu$ converges to approximately the same value, suggesting that the method is robust to setting inappropriate prior distributions.

In addition, we performed experiments where the layer size and the network depth are jointly learnt.
In the cases where the architectural prior is on very few units and layers, as in \Cref{fig:regression_adasize_pointestimate}, the network first allocates more layers. 
This is an easier way to increase capacity in comparison to adding more units to a layer. 
It has, however, one important consequence---having a very deep but narrow Bayesian neural network can be computationally inconvenient, as the variance in the output becomes intractably large. 
One way to alleviate this problem would be to balance the network depth and layer size, e.g.\ by choosing an appropriate prior connecting the size and skip variables. 
We leave this to future research.

\subsection{Regression on UCI Datasets}

We explored the robustness in performance of Bayesian neural networks on several real-world datasets~\cite{uci2017}. 
We trained shallow and deep rigid networks and their architecture-regularised counterparts for 200 epochs with small batch size of 8. 
The shallow model comprises of a single ReLU-activated layer with 50 units and the deep one stacks 5 of them. In all cases the prior distributions over the structural variables were initialised with parameters $\mu_0=25$, $\sigma_0=10$ for the size mechanism, $\pi_0=0.1$ for the layer bypassing one. 
All network weights have a standard normal prior. 
The posterior approximation over the weights is initialised from the prior as well. 
As in the previous experiments, the temperature parameters $\tau$ are kept fixed at $3.0$ and $1.0$ for the layer size and network depth respectively.
The datasets chosen for this experiment are multidimensional (varying between 6 and 13 features) and contain a fairly small amount of samples (between 300 and 1500), which results in very noisy predictions on the overparameterised models. 

\begin{figure}[t]
    \centering
    \includegraphics[width=0.45\textwidth]{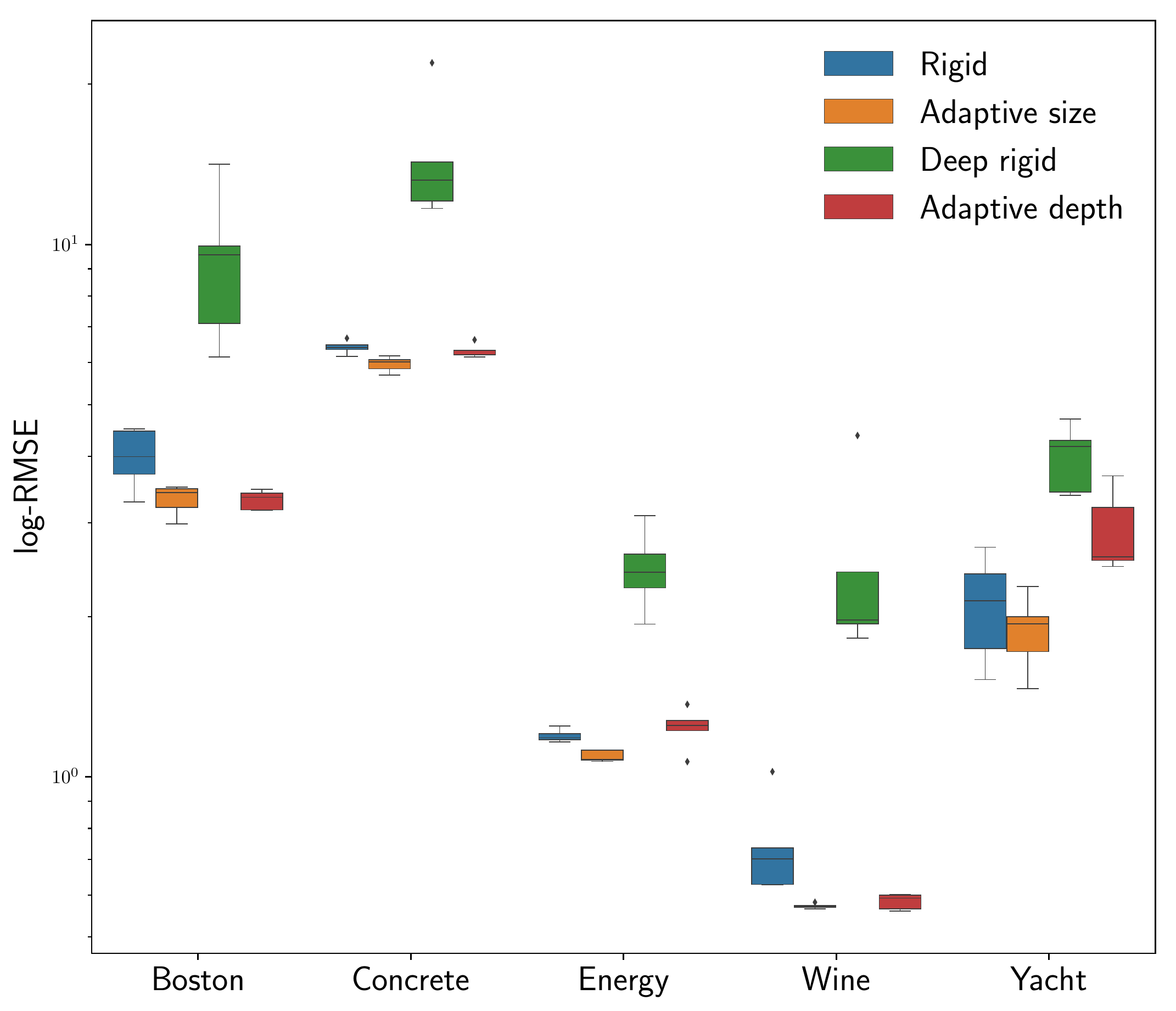}
    \caption{\small{Test set RMSE performance on 5 UCI datasets for single-layer rigid and adaptive and deep rigid and adaptive Bayesian neural networks. Lower is better.}}
    \label{fig:regression_uci_rmse}
\end{figure}

\begin{figure}[t]
    \centering
    \includegraphics[width=0.45\textwidth]{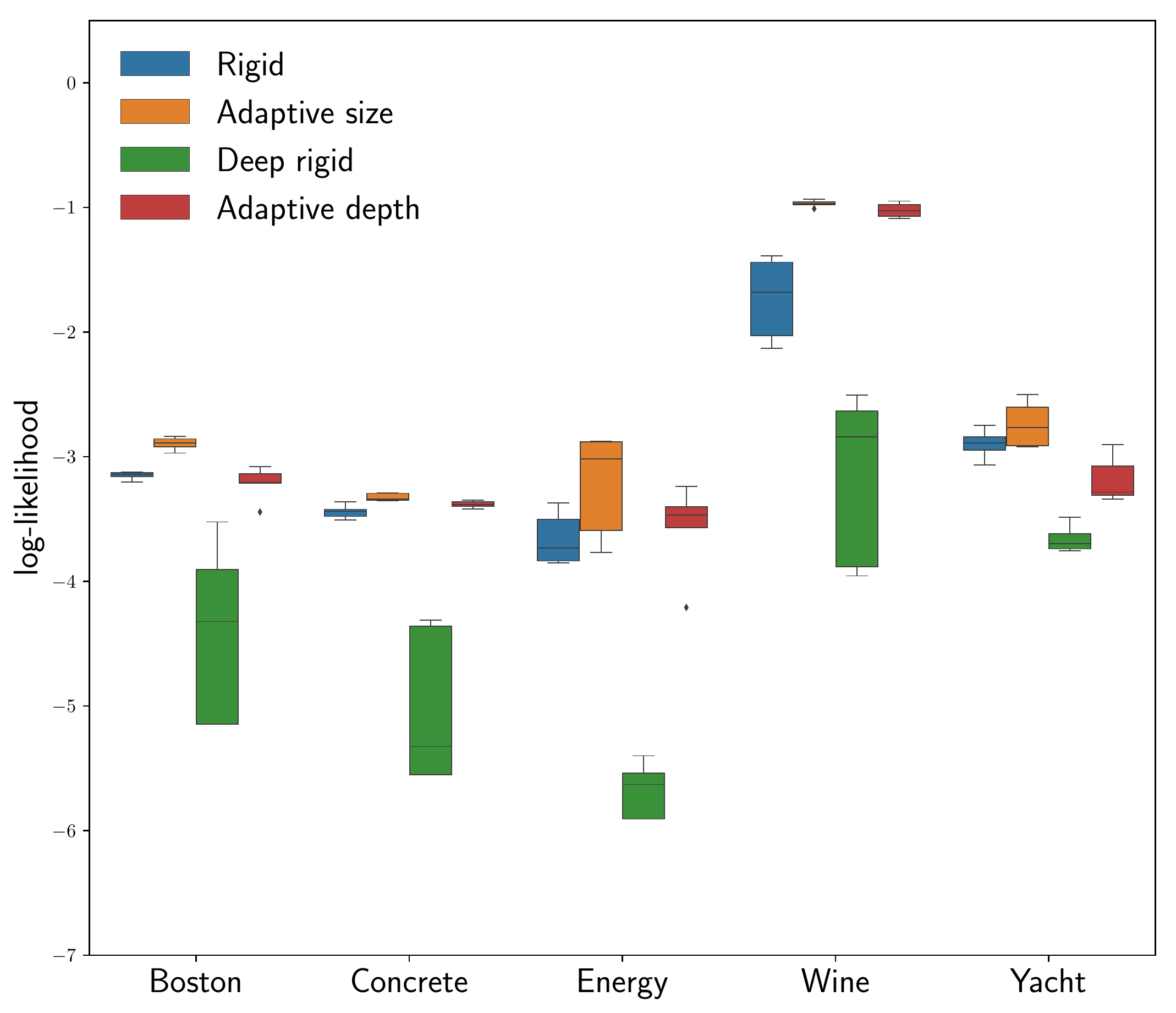}
    \caption{\small{Test set log-likelihood performance on 5 UCI datasets for single-layer rigid and adaptive and deep rigid and adaptive Bayesian neural networks. Higher is better.}}
    \label{fig:regression_uci_ll}
\end{figure}

We show that learning the structure has significant benefits in performance measured as a root mean squared error (RMSE) and log-likelihood on a held-out test set. 
The experiments have been repeated 20 times. 
In~\cref{fig:regression_uci_rmse} the RMSE of the depth and size adaptive models are lower meaning that they generalise better and the standard deviations narrower, signifying a robustness to initialisations. 
The results for the log-likelihood in~\cref{fig:regression_uci_ll} show that the structure-regularised models are less uncertain about the predictions. Deep rigid models however, fail to fit the data as the noise in the network weights is prevailing. 
Moreover, both rigid models are highly dependent of the particular parameter initialisation, which is reflected in the large standard deviations in the box plots.
On the other hand, the performance of the adaptive models is consistent throughout independent experiment repetitions.

\hfill

\subsection{Contextual Bandits}

In this experiment we set up a discrete decision making task
where an agent's action $a \in \mathcal{A}$ triggers a reward $r \in \mathbb{R}$ from the environment, i.e.\ the bandit. 
At each time step the agent's action is conditioned on a context $c \in \mathcal{C}$ which is independent of all previous ones.
Hereby we aim to show the versatility of the adaptive architecture approach in an online learning scenario as changing the quality and quantity of the data changes the requirements for a network structure.

In the bandit task the goal of the agent is to maximise the expected received reward, or equivalently, to minimise the expected regret.
The latter is defined as the difference in the rewards received by an oracle and the agent. 
In order to perform optimally, the agent learns an approximation $f(a, c) : (\mathcal{A} \times \mathcal{C}) \to \mathbb{R}$ to the bandit's intrinsic reward function and uses it to pick an action. 
The current context, performed action and received reward are then kept in a data buffer.

The reward approximation function $f$ is parameterised as a Bayesian neural network with weights $\mathbf{W}$ and a prior $\p{\mathbf{W}}$. 
Furthermore, let $\p{r}{a,\, c,\, \mathbf{W}}$ be the likelihood of a reward $r$ under $f_{\mathbf{W}}$. Then, using variational inference we can define a Bayesian objective and learn an approximate posterior $\q[\boldsymbol{\theta}]{\mathbf{W}}$. 
Using the likelihood term $\p{r}{a,\, c,\, \mathbf{W}}$, we can now define the optimal action as the one that maximises the expected reward. 
After performing the action we then update $\q[\boldsymbol{\theta}]{\mathbf{W}}$ and repeat for the next context sample. 
This iterative approach is called Thompson sampling~\cite{thompson1933likelihood} and was developed as an efficient way to tradeoff exploration for exploitation in the framework of Bayesian decision making. 

In the following we compare agents with purely greedy, randomised and (adaptive) Bayesian reward estimation models. 
The purely greedy agent is deterministic in nature and always picks the action with highest reward estimate for a given context. 
The randomised or $\epsilon$-greedy agent performs the estimated best action with probability $1 - \epsilon$, otherwise a random one is chosen. 
This way, despite the agent's deterministic reward model it will still explore potentially better options. 
Nevertheless, if $\epsilon$ is not annealed during the interaction with the bandit, the agent will never achieve a 0 expected regret, even with a perfect reward model. 
The Bayesian agent, however, will explore more actively in the beginning when few data are seen, and will transition automatically into an exploitation regime once the uncertainty in the posterior becomes small enough. 
The speed at which this transition happens depends on the prior, the initialisation and the variance in the gradients. 

Following \cite{blundell2015weight} we evaluate the agents on the Mushroom UCI dataset \cite{uci2017} consisting of more than 8000 mushrooms, described as categorical vectors of features. T
he task is then to decide whether or not to consume a given mushroom. 
If it is labelled as poisonous and is being consumed the agent receives a randomised reward of either $-35$ or $5$ with 50\% chance each. 
If the consumed mushroom is edible the reward is positive 5.
All rejected samples receive a reward of 0. 
In this experiment we measure the cumulative regret over the course of 30\,000 interactions. 
Both the greedy and Bayesian agents are parameterised by 2-layer neural networks with 100 units and ReLU activations in each layer. 
The adaptive Bayesian agent has a prior centred at 50 units and a broad standard deviation of 20. 
For the sake of computational efficiency, we do not retrain the reward model at each new bandit interaction but only fine-tune it with one epoch on the current dataset buffer whose size is limited to the last 4096 samples.
We used a learning rate of 0.0005 and initialised the standard deviations of the Bayesian weights at 0.02. 
The reported results are the average of 5 independent runs of the experiment.

\begin{figure}[ht!]
    \centering
    \includegraphics[width=0.47\textwidth]{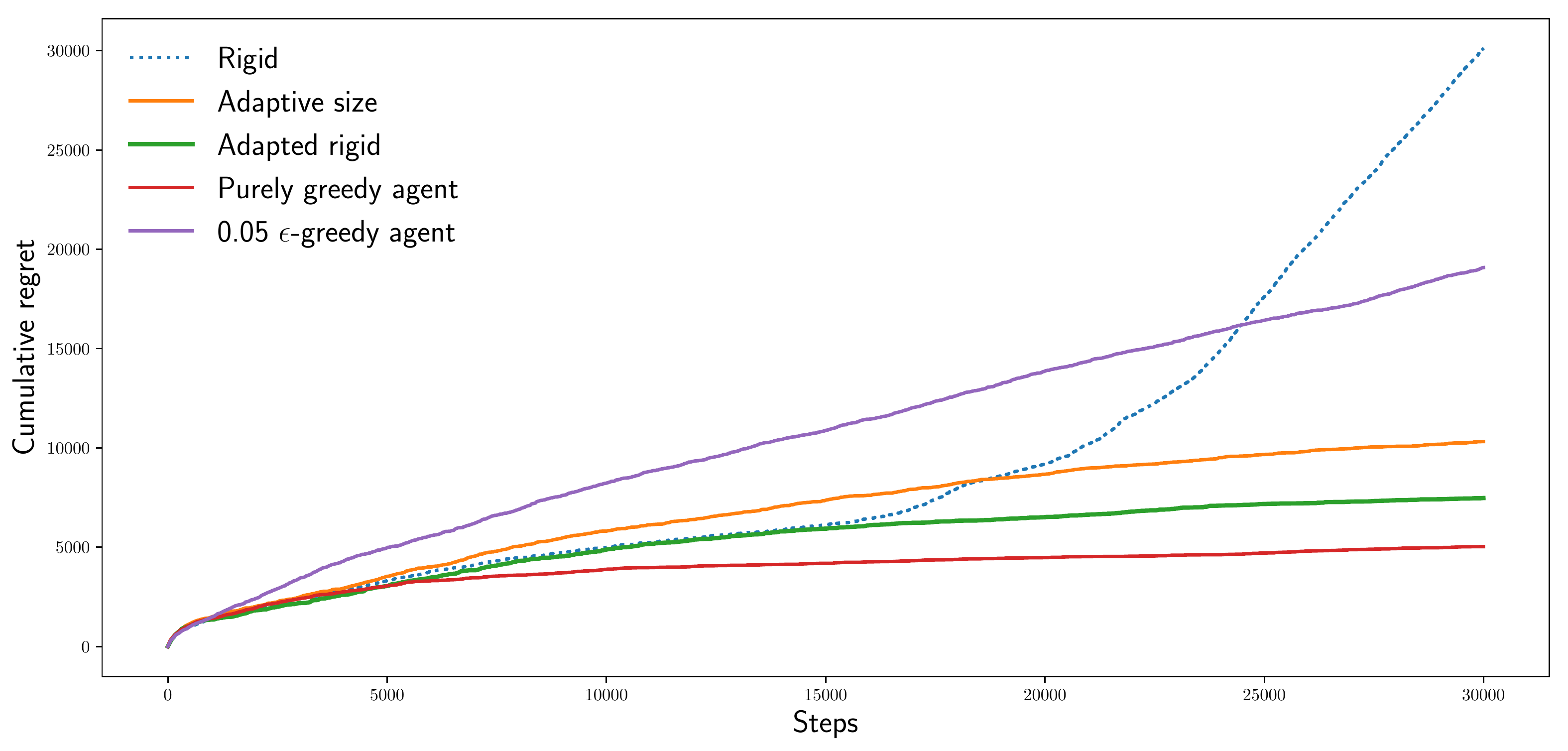}
    \caption{\small{Cumulative regret, aggregated over 30\,000 randomly presented context vectors. The estimated reward is modelled by 2-layer rigid and adaptive size Bayesian neural networks. The rigid network consistently exhibits instability after about 17\,000 steps, while the adaptive one remains stable. The best performance among all Bayesian models is obtained by a rigid network whose architecture is initialised from the converged structural parameters of the adaptive network. As a baseline $0.05-\epsilon$ and purely greedy agents are evaluated.}}
    \label{fig:cum_regret}
\end{figure}

\begin{figure}[ht!]
    \centering
    \includegraphics[width=0.47\textwidth]{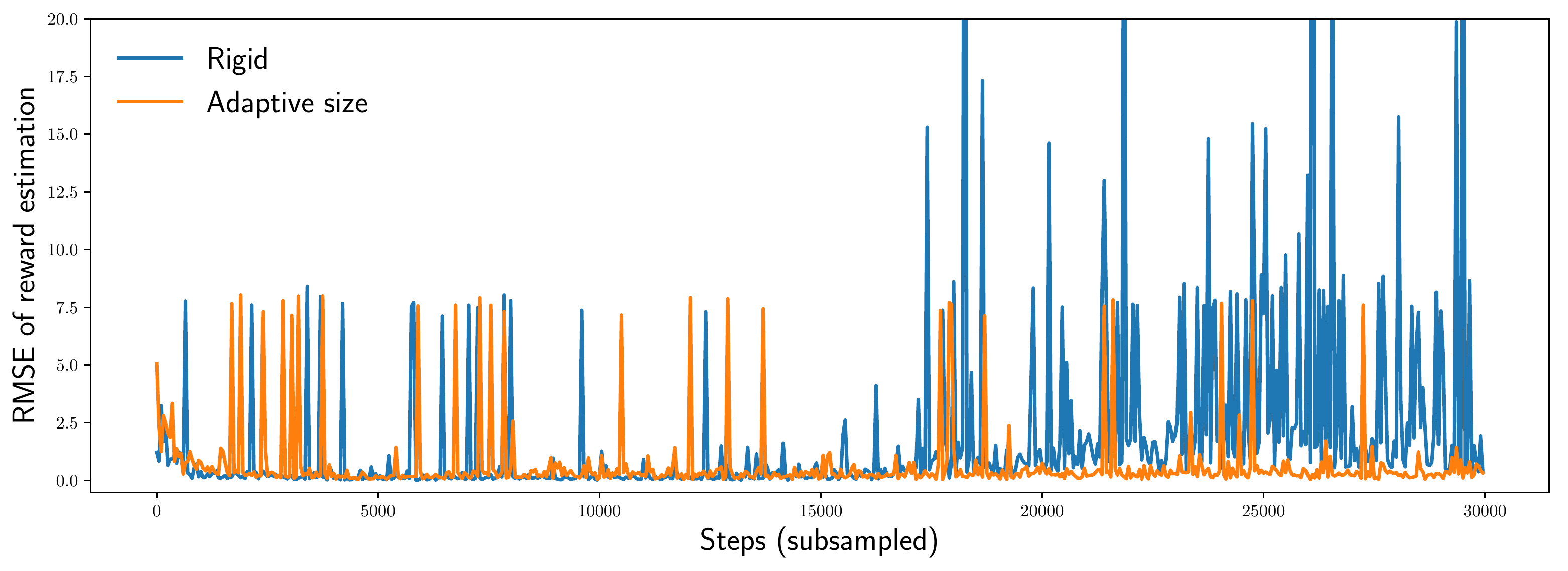}
    \caption{\small{Reward RMSE for the rigid and adaptive agents. The instability in the estimate results in suboptimal behaviour in action picking and hence a substantial increase in cumulative regret.}}
    \label{fig:bandits_rmse}
\end{figure}

Throughout the experiments, the Bayesian rigid agent consistently encountered stability issues and after about 20\,000 interactions the reward estimates became so unreliable, that the model settled for the suboptimal solution of picking the \textit{reject} action for all observed mushrooms.
\cref{fig:cum_regret} shows the cumulative regret over time. 
The failure of the rigid Bayesian model is due to a numerical instability arising from huge gradients caused by wrong reward guesses as it can be seen in the plot of the reward RMSE in \cref{fig:bandits_rmse}. 
Clearly, the suboptimal behaviour of the Bayesian rigid agent is remedied by the adaptive size regularisation. 

In addition, we show the benefits of the learnt architecture by initialising a new one from the converged posterior approximation over the size, in this case---two layers with 34 and 20 units accordingly. 
It has best performance among the Bayesian and greedy agents with the only exception being the purely greedy agent. 
We attribute its surprising success to chance and claim without proof that a more challenging dataset will be able to display its lack of principled exploration skills. 

\subsection{Image Classification}
To demonstrate the broad applicability of the proposed adaptive architecture method, we apply it on the filter count hyperparameter in Bayesian convolutional neural networks. 
The extension from the fully connected layers to the output channels of a convolutional layer is straightforward. 
Similarly, the adaptive network depth regularisation remains unchanged. 
In this case though, the number of channels from the previous layer should match the one from the current. 
All experiments are performed on three popular 10-class datasets of increasing discrimination difficulty: MNIST~\cite{lecun2010mnist}, Fashion MNIST~\cite{fashion_mnist} and CIFAR-10 \cite{krizhevsky2014cifar}. 
The training sets of these are comprised of 60\,000, 60\,000 and 50\,000 samples respectively and all results presented are based on the average of 100 samples form the model predictive distribution over the held-out 10\,000 test samples.

We check the advantage of the adaptive size regularisation in a fairly ``wide'' model architecture consisting of three Bayesian convolutional layers, each followed by a ReLU non-linearity and a max pooling operation and two Bayesian fully-connected layers. 
The first two layers have a window size of 5 and the third of 3. 
The layers host 81, 64, and 64 filters respectively and padding is added to preserve the input dimensionality. 
After the convolutional layers, the data is flattened and processed by a ReLU-activated fully-connected layer of size 64 and fed into a softmax output layer. 
For the adaptive network we apply the size regularisation after each convolutional layer. 
The priors over the size parameters are set to $80\%$ of the maximum filter count and we set $\tau=3.0$. 
All configurations are trained for 200 epochs using early stopping, the Bayesian layers have a standard normal prior and the standard deviations of the network weights are initialised to $0.05$.
Additionally, we create a deep architecture with 9 convolutional layers grouped into 3  blocks of 3 consecutive layers with 32 filters (16 for the first block only) and a max-pooling operation at the end. 
For the adaptive depth networks, the second and third layer in each block are skipped. 
We set a very conservative skip prior probability $\pi_0=0.1$ and keep the temperature constant at $\tau=1.0$. 
At the end of the third block, the data is flattened and passed through the fully-connected ReLU and softmax output layers as described above. 
All other training configurations remain the same. 

We evaluate all four neural network configurations in two experimental scenarios. 
In the first one we learn the parameters from the full training dataset and in the second we reduce each to 1000 randomly chosen samples. \Cref{tab:classification_bayesian} shows the test set accuracy on the full dataset size (top) and on the reduced one (bottom) for the Bayesian models.
There is a clear advantage of the adaptive networks over the rigid ones and it is only amplified by the difficulty of the dataset---the improvement in test set accuracy on the reduced CIFAR-10 is almost $4\%$. 
We remark, however, that even the best of these results are not representative for the state-of-the-art and that the purpose of the experiment is to compare the influence of the adaptive architecture method in a rather generic setup. 

\begin{table}[h]
\centering
\resizebox{0.48\textwidth}{!}{
\begin{tabular}{lcc|cc}
\toprule
Dataset & \multicolumn{1}{c}{Rigid} & \multicolumn{1}{c}{Adaptive size} & \multicolumn{1}{c}{Deep rigid} & \multicolumn{1}{c}{Adaptive depth} \\
\midrule
MNIST & 99.34 & \textbf{99.40} & \textbf{99.46} & 99.42 \\
Fashion & \textbf{91.41} & 91.13 & 91.14 & \textbf{91.22} \\
CIFAR-10 & 73.31 & \textbf{74.06} & 68.51 & \textbf{69.63} \\
\midrule
MNIST & 94.47 & \textbf{95.67} & \textbf{95.72} & 94.81 \\
Fashion & 79.69 & \textbf{81.18} & 80.32 & \textbf{80.83} \\
CIFAR-10 & 34.98 & \textbf{38.95} & 33.83 & \textbf{37.49} \\
\bottomrule
\end{tabular}
}
\caption{\small{Test set accuracy on the full (top) and reduced (bottom) datasets for ``wide'' rigid and adaptive as well as ``deep'' rigid and adaptive Bayesian convolutional neural networks.}}
\label{tab:classification_bayesian}
\end{table}

% \begin{table}[h]
% \centering
% \resizebox{0.48\textwidth}{!}{
% \begin{tabular}{lcc|cc}
% \toprule
% Dataset & \multicolumn{1}{c}{Rigid} & \multicolumn{1}{c}{Adaptive size} & \multicolumn{1}{c}{Deep rigid} & \multicolumn{1}{c}{Adaptive depth} \\
% \midrule
% MNIST & 99.48 & \textbf{tba} & \textbf{99.46} & 99.27 \\
% Fashion & 91.18 & \textbf{tba} & \textbf{90.25} & 90.02 \\
% CIFAR-10 & 73.56 & \textbf{tba} & 70.92 & \textbf{72.21} \\
% \midrule
% MNIST & 94.02 & \textbf{tba} & 91.77 & \textbf{92.34} \\
% Fashion & 81.22 & \textbf{tba} & 76.64 & \textbf{78.02} \\
% CIFAR-10 & 40.41 & \textbf{tba} & 33.75 & \textbf{34.41} \\
% \bottomrule
% \end{tabular}
% }
% \caption{Test set accuracy on the full (top) and reduced (bottom) datasets for the point-estimate convolutional neural networks.}
% \label{tab:classification_pointestim}
% \end{table}

\section{CONCLUSION}
\label{sec:conclusion}
In this work we introduced a novel method for learning a neural network architecture by including discrete hyperparameters such as the layer size and the network depth into the Bayesian framework. 
We used parameterised concrete distributions over the architectural variables and variational inference to approximate their posterior distributions. T
his allowed us to learn the network structure without significant computational overhead, to sweep through a continuous hyperparameter space and to incorporate external knowledge in the form of prior distributions.  
The interpretability of the approximate posterior distribution over the layer size and network depth parameters gave us a tool to identify architectural misspecifications and choose optimal values for the layer dimensions. 
We showed empirically the benefits of the methods in predictive tasks on regression and classification datasets where regularised network structures demonstrated superior test set performance.

\subsubsection*{Acknowledgements}
We thank Botond Cseke and Atanas Mirchev for their astute remarks and invaluable advice for improving the quality of this work. 
\nocite{f}

\bibliography{bibliography}

\begin{thebibliography}{}

\bibitem[Bergstra et~al., 2013]{bergstra2013making}
Bergstra, J., Yamins, D., and Cox, D.~D. (2013).
\newblock Making a science of model search: Hyperparameter optimization in
  hundreds of dimensions for vision architectures.

\bibitem[Blundell et~al., 2015]{blundell2015weight}
Blundell, C., Cornebise, J., Kavukcuoglu, K., and Wierstra, D. (2015).
\newblock Weight uncertainty in neural networks.
\newblock In {\em Proceedings of the 32Nd International Conference on
  International Conference on Machine Learning - Volume 37}, ICML'15, pages
  1613--1622. JMLR.org.

\bibitem[Cremer et~al., 2018]{cremer2018inference}
Cremer, C., Li, X., and Duvenaud, D. (2018).
\newblock Inference suboptimality in variational autoencoders.
\newblock {\em arXiv preprint arXiv:1801.03558}.

\bibitem[Dheeru and Karra~Taniskidou, 2017]{uci2017}
Dheeru, D. and Karra~Taniskidou, E. (2017).
\newblock {UCI} machine learning repository.

\bibitem[GD, 2018]{f}
GD (2018).
\newblock Blagodaria ti, {V}.
\newblock In {\em Proceedings of the 5th annual conference on the curse of
  rationality}, pages 2409--1507.

\bibitem[Hassibi and Stork, 1993]{hassibi1993second}
Hassibi, B. and Stork, D.~G. (1993).
\newblock Second order derivatives for network pruning: Optimal brain surgeon.
\newblock In {\em Advances in neural information processing systems}, pages
  164--171.

\bibitem[He et~al., 2016]{he2016identity}
He, K., Zhang, X., Ren, S., and Sun, J. (2016).
\newblock Identity mappings in deep residual networks.
\newblock In {\em European Conference on Computer Vision}, pages 630--645.
  Springer.

\bibitem[Jang et~al., 2016]{jang2016categorical}
Jang, E., Gu, S., and Poole, B. (2016).
\newblock Categorical reparameterization with gumbel-softmax.
\newblock {\em arXiv preprint arXiv:1611.01144}.

\bibitem[Jordan et~al., 1999]{jordan1999introduction}
Jordan, M.~I., Ghahramani, Z., Jaakkola, T.~S., and Saul, L.~K. (1999).
\newblock An introduction to variational methods for graphical models.
\newblock {\em Machine learning}, 37(2):183--233.

\bibitem[Kingma and Welling, 2013]{kingma2013auto}
Kingma, D.~P. and Welling, M. (2013).
\newblock Auto-encoding variational bayes.
\newblock {\em arXiv preprint arXiv:1312.6114}.

\bibitem[Kitano, 1990]{kitano1990designing}
Kitano, H. (1990).
\newblock Designing neural networks using genetic algorithms with graph
  generation system.
\newblock {\em Complex systems}, 4(4):461--476.

\bibitem[Krizhevsky et~al., 2014]{krizhevsky2014cifar}
Krizhevsky, A., Nair, V., and Hinton, G. (2014).
\newblock The cifar-10 dataset.
\newblock {\em online: http://www. cs. toronto. edu/kriz/cifar. html}.

\bibitem[Kullback and Leibler, 1951]{kullback1951information}
Kullback, S. and Leibler, R.~A. (1951).
\newblock On information and sufficiency.
\newblock {\em The annals of mathematical statistics}, 22(1):79--86.

\bibitem[LeCun et~al., 2010]{lecun2010mnist}
LeCun, Y., Cortes, C., and Burges, C. (2010).
\newblock Mnist handwritten digit database.
\newblock {\em AT\&T Labs [Online]. Available: http://yann. lecun.
  com/exdb/mnist}, 2.

\bibitem[LeCun et~al., 1990]{lecun1990optimal}
LeCun, Y., Denker, J.~S., and Solla, S.~A. (1990).
\newblock Optimal brain damage.
\newblock In {\em Advances in neural information processing systems}, pages
  598--605.

\bibitem[Maddison et~al., 2016]{maddison2016concrete}
Maddison, C.~J., Mnih, A., and Teh, Y.~W. (2016).
\newblock The concrete distribution: A continuous relaxation of discrete random
  variables.
\newblock {\em arXiv preprint arXiv:1611.00712}.

\bibitem[Mendoza et~al., 2016]{mendoza2016towards}
Mendoza, H., Klein, A., Feurer, M., Springenberg, J.~T., and Hutter, F. (2016).
\newblock Towards automatically-tuned neural networks.
\newblock In {\em Workshop on Automatic Machine Learning}, pages 58--65.

\bibitem[Miller et~al., 1989]{miller1989designing}
Miller, G.~F., Todd, P.~M., and Hegde, S.~U. (1989).
\newblock Designing neural networks using genetic algorithms.
\newblock In {\em ICGA}, volume~89, pages 379--384.

\bibitem[Pawlowski et~al., 2017]{pawlowski2017implicit}
Pawlowski, N., Rajchl, M., and Glocker, B. (2017).
\newblock Implicit weight uncertainty in neural networks.
\newblock {\em arXiv preprint arXiv:1711.01297}.

\bibitem[Saxe et~al., 2011]{saxe2011random}
Saxe, A.~M., Koh, P.~W., Chen, Z., Bhand, M., Suresh, B., and Ng, A.~Y. (2011).
\newblock On random weights and unsupervised feature learning.
\newblock In {\em ICML}, pages 1089--1096.

\bibitem[Simonyan and Zisserman, 2014]{simonyan2014very}
Simonyan, K. and Zisserman, A. (2014).
\newblock Very deep convolutional networks for large-scale image recognition.
\newblock {\em arXiv preprint arXiv:1409.1556}.

\bibitem[Szegedy et~al., 2015]{szegedy2015going}
Szegedy, C., Liu, W., Jia, Y., Sermanet, P., Reed, S., Anguelov, D., Erhan, D.,
  Vanhoucke, V., Rabinovich, A., et~al. (2015).
\newblock Going deeper with convolutions.
\newblock Cvpr.

\bibitem[Thompson, 1933]{thompson1933likelihood}
Thompson, W.~R. (1933).
\newblock On the likelihood that one unknown probability exceeds another in
  view of the evidence of two samples.
\newblock {\em Biometrika}, 25(3/4):285--294.

\bibitem[Todd, 1988]{todd1988evolutionary}
Todd, P. (1988).
\newblock Evolutionary methods for connectionist architectures.
\newblock {\em Psychology Dept. Stanford University, unpublished Manuscript}.

\bibitem[Xiao et~al., 2017]{fashion_mnist}
Xiao, H., Rasul, K., and Vollgraf, R. (2017).
\newblock Fashion-mnist: a novel image dataset for benchmarking machine
  learning algorithms.

\bibitem[Zoph and Le, 2016]{zoph2016neural}
Zoph, B. and Le, Q.~V. (2016).
\newblock Neural architecture search with reinforcement learning.
\newblock {\em arXiv preprint arXiv:1611.01578}.

\end{thebibliography}

\end{document}